\documentclass[letterpaper, 10 pt, conference]{ieeeconf}

\IEEEoverridecommandlockouts

\usepackage{graphicx}

\usepackage{amsmath}
\usepackage{amssymb}
\usepackage{hyperref}
\usepackage{cleveref}
\crefname{figure}{Fig.}{Figs.}
\Crefname{figure}{Fig.}{Figs.}
\crefname{table}{Table}{Tables}
\Crefname{table}{Table}{Tables}
\usepackage{tabularx}
\usepackage{booktabs}

\newcommand{\papertablestyle}{%
  \footnotesize
  \setlength{\tabcolsep}{3pt}%
  \renewcommand{\arraystretch}{1.06}%
}
\newcommand{\twolinetablecell}[2]{\shortstack{#1\\[-1pt]#2}}

\title{\LARGE \bf
Embodiment-Aware Inference-Time Search for Human\\Reaction Selection in Humanoid Retargeting
}

\author{
Yongpeng Cao$^{1, *}$\thanks{$^{*}$Corresponding author}%
\thanks{$^{1}$Institute of Industrial Science, The University of Tokyo, Japan}
, Masahiro Hirano$^{1}$, and Yuji Yamakawa$^{2}$%
\thanks{$^{2}$Interfaculty Initiative in Information Studies, Graduate School of Interdisciplinary Information Studies, The University of Tokyo, Japan}%
\\ \{cao, mhirano, y-ymkw\}@iis.u-tokyo.ac.jp
}

\begin{document}
\bstctlcite{IEEEref:BSTcontrol}

\maketitle
\thispagestyle{empty}
\pagestyle{empty}

\begin{abstract}
Training humanoid robots to learn human behaviors requires expensive data collection, and human--robot interaction data can be even more difficult and costly to collect. Training a specialized reaction model for each humanoid embodiment can also be time-consuming and resource-intensive. Human-to-humanoid motion retargeting provides a more scalable alternative, allowing human--human reaction data to serve as a common motion representation for different humanoid embodiments. However, due to morphological and kinematic differences between humans and humanoid robots, a reaction preferred in human motion space may not be the most suitable one for execution on a particular robot embodiment.

In this work, we propose an inference-time reaction search framework that leverages stochastic sampling from a generative reaction model and jointly considers reaction quality and humanoid embodiment feasibility. Given the observed interaction, our framework generates multiple plausible reaction candidates. Each candidate is retargeted to the target humanoid and evaluated using complementary interaction and embodiment-specific feasibility criteria, allowing candidate selection under the physical constraints of the target embodiment. Since the reaction generator remains frozen, the framework can adapt to different humanoid embodiments at inference time without retraining. Experimental results demonstrate that the proposed framework maintains overall human reaction quality while improving the feasibility of retargeted humanoid reactions across different embodiments. The proposed pipeline improves humanoid retargeting feasibility by \(1.61\%\). Moreover, human-space and G1 best-candidate selections disagree in \(78.98\%\) of samples, demonstrating substantial embodiment dependence. Code will be released at \url{https://anonymous.4open.science/r/Runtime-Humanoid-Reaction-Searching-2BF2}.
\end{abstract}

\section{Introduction}

As humanoid robots move beyond structured industrial settings into human-centered environments, the ability to interact naturally and responsively with people becomes a key challenge. To achieve this, humanoids must be able to generate plausible reactions to human actions. However, training humanoids to learn human behaviors requires expensive data collection, while human--robot interaction data can be even more complex and costly to collect. Training a specialized reaction model for each humanoid embodiment can therefore be time-consuming and resource-intensive.

Human motion data, including reactions collected from human--human interactions, are comparatively more accessible and scalable to collect and have been used as motion references for humanoid learning. Existing works leverage human motion data to train humanoid behaviors by adapting human motions to the target humanoid and subsequently learning control policies from the resulting motions~\cite{fu2024humanplus, videomimic}. Human motion can therefore serve as a common intermediate representation that can be transferred to different humanoid embodiments.

However, due to morphological and kinematic differences between the human body and humanoid robots, a motion that is plausible as a human reaction may not necessarily be the most suitable one for retargeting and execution on a particular embodiment. Consequently, the reaction preferred in human-motion space and the reaction most suitable after humanoid retargeting may correspond to different candidates. For example, differences in limb proportions and joint ranges can make a human reaction easy to reproduce on one humanoid but difficult on another. Different humanoid embodiments may therefore prefer different reaction candidates according to their embodiment-specific constraints.

Human reactions are inherently one-to-many: given the same observed interaction, multiple future reactions can be plausible. Stochastic generative models can capture these plausible outcomes and produce diverse reaction candidates through inference-time sampling. We exploit this capability by sampling multiple reaction candidates from a frozen generator and evaluating them after retargeting according to the requirements of the target embodiment. Nevertheless, inference-time selection among reaction candidates based on their suitability after humanoid retargeting remains relatively underexplored.

In this work, we propose an inference-time reaction search framework that jointly considers human reaction quality and humanoid embodiment feasibility. Given an observed interaction, our framework generates multiple causal reaction candidates, retargets each candidate to the target humanoid, and evaluates them using complementary interaction-preservation and embodiment-specific feasibility criteria. Rather than treating retargeting as a downstream post-processing step, we incorporate post-retargeting evaluation into reaction selection. Experimental results demonstrate that the proposed framework maintains overall human-reaction quality while improving the feasibility of retargeted humanoid reactions across different embodiments, including a \(1.61\%\) retargeting feasibility gain on the Inter-X test set.

Our main contributions are:
\begin{itemize}
\item We formulate embodiment-aware reaction selection, showing that the preferred human reaction may differ from the candidate most suitable after humanoid retargeting.
\item We propose a causal inference-time search framework that samples multiple candidate reactions from a frozen generator and selects them after retargeting using interaction and embodiment-feasibility criteria.
\item We validate the framework across different retargeters, interaction datasets, and humanoid embodiments, demonstrating consistent gains in humanoid retargeting feasibility.

\end{itemize}

\section{Related Work}
\subsection{Human Interaction and Reaction Generation}

Human--human reaction generation has received increasing attention as a critical extension of human interaction synthesis, where the interaction of multiple people must remain spatially, temporally, and semantically coordinated. Recent approaches such as InterGen~\cite{liang2024intergen} and InterMask~\cite{ICLR2025_intermask} explore joint interaction generation, including reactions, using DiT and VQ-VAE architectures. Human--human interaction datasets such as InterHuman~\cite{liang2024intergen} and Inter-X~\cite{xu2024interx} further provide diverse motions and benchmarks for learning and evaluating interactive motion generation.

A topic more closely related to our work is action-to-reaction generation, where the motion of one agent is generated conditionally on the motion of another. InterFormer~\cite{chopin2023interformer} introduces an interaction-aware formulation for human reaction generation. ReGenNet~\cite{xu2024regennet} further achieves online reaction generation without access to future motion. Recent methods such as Think-Then-React~\cite{tan2025thinkthenreact} introduce LLM-based semantic reasoning to generate appropriate reactions. Ready-to-React~\cite{cen2025_ready_to_react}, ARMFlow~\cite{geng2026armflow}, and ReMoGen~\cite{ye2026remogen} explore real-time online reaction synthesis using efficient models and training strategies. Human-X~\cite{ji2025humanx} extends reaction generation with physically plausible humanoid reaction synthesis.

Despite these advances in model architectures and training strategies, most existing methods adopt a fixed inference procedure. Comparatively less attention has been paid to inference-time search to refine the generated reactions. In this work, we instead take the pretrained reaction generator as a proposal model and select among multiple candidate reactions based on embodiment requirements.

\subsection{Human-to-Humanoid Motion Retargeting}

Learning generalizable humanoid behaviors typically requires large and diverse motion datasets. Collecting robot-native motion trajectories at such scale, however, remains substantially more expensive than obtaining human motion data. Consequently, retargeting existing human motion data to humanoids has become a labor-saving and promising approach for constructing training references for humanoid control.

Existing retargeting methods address the human--humanoid embodiment gap using optimization-based and learning-based hybrid approaches. GMR~\cite{joao2025gmr} converts human motions into robot joint configurations while explicitly considering the target humanoid morphology. NMR~\cite{zhao2026nmr} instead learns the temporal human-to-humanoid mapping with optimization-based methods to reduce artifacts. Other methods such as CoRe~\cite{CoRe2025} and Human2Humanoid~\cite{huang2026human2humanoid} further incorporate morphology constraints, contact, and unified representations to improve cross-embodiment transfer.

Following retargeting, physics-based tracking frameworks such as ProtoMotions~\cite{ProtoMotions} and TWIST~\cite{ze2025twist} use retargeted human motion as reference trajectories to learn physically valid humanoid behaviors. Interactive systems such as Human-X also condition the tracking policy on the observed actor to adapt to deviations in the actual interaction.

These approaches primarily address how a given human reference should be transferred to a target humanoid. However, when there are multiple possible reactions and a reaction generator can provide multiple plausible human reactions, the choice of the reference reaction itself becomes an additional degree of freedom. Our work therefore considers reaction selection jointly with humanoid embodiment compatibility rather than only retargeting a predetermined reaction. This formulation allows the system to select a reaction that simultaneously preserves the intended interaction and provides a more favorable reference for humanoid execution.

\subsection{Inference-Time Search and Embodiment-Aware Motion Selection}

Inference-time sampling and optimization complement learning-based methods to improve model performance. In human motion generation, DNO~\cite{karunratanakul2024dno} optimizes the noise input of a frozen diffusion model based on task-specific objectives, while PhysDiff~\cite{yuan2023physdiff} guides the diffusion process by introducing physics-based projection to adapt to practical scenarios. Diffusion-ES~\cite{yang2024diffusionES} uses generated trajectory proposals for repeated evaluation and refinement, and sampling-based methods evaluate candidate trajectories according to related task costs. PointWorld~\cite{Huang_2026_pointworld} similarly evaluates future states to select and refine robot action candidates. Recent humanoid control frameworks such as BeyondMimic~\cite{liao2026beyondmimic} further demonstrate that task-specific guidance can be applied at inference time to adapt motion behaviors to different tasks without retraining the generation model for each new task.

These methods demonstrate that additional inference-time computation can improve the task suitability of motions. Reaction generation allows multiple plausible reactions, while retargeting imposes embodiment-dependent constraints on which reactions are suitable for robot execution. However, existing motion-search and guidance methods generally optimize motions directly in human or task space, while humanoid retargeting approaches usually assume that the human reference trajectory has already been selected. The integration of these two components remains underexplored. Our work addresses this gap through an inference-time search framework that samples multiple causal human reactions from a frozen generator, retargets each candidate to the target humanoid, and selects among them using robot-side kinematic and interaction criteria. This formulation allows reaction selection to depend on the target embodiment while leaving the human reaction generator unchanged.

\section{Methodology}

\begin{figure*}[htbp!]
\vspace*{5pt}
\centering
\includegraphics[width=\textwidth]{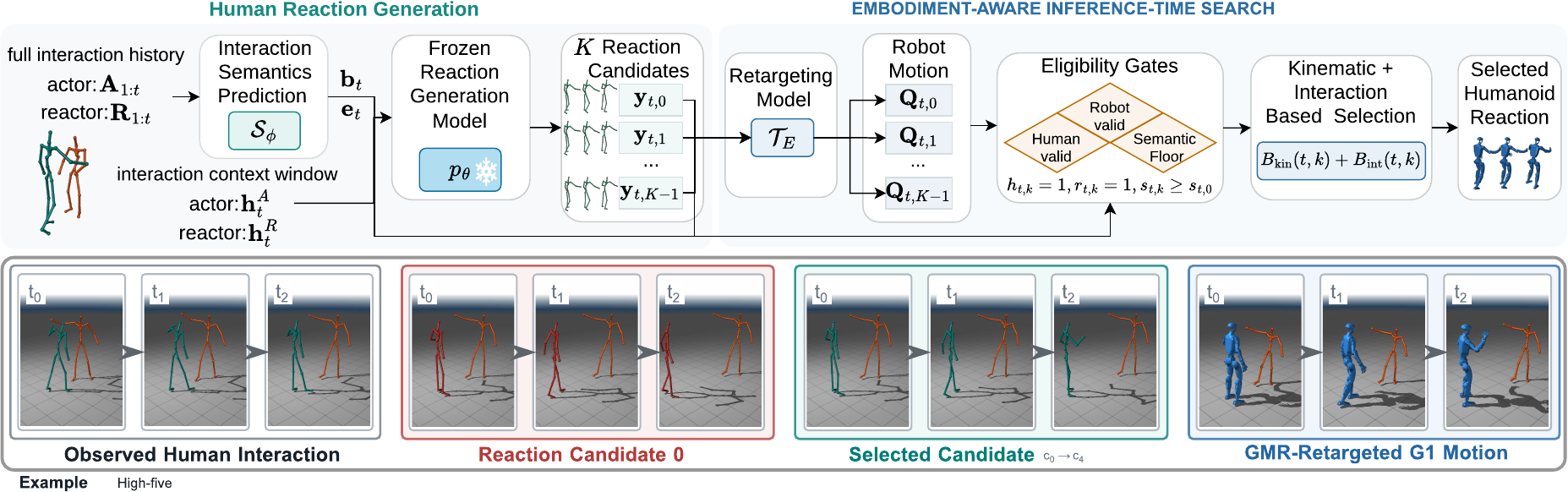}
\caption{Overview of the proposed inference-time search framework. An interaction semantic condition is generated using the observed interaction history for the frozen ReMoGen proposal generator. Each of the $K$ reactions is retargeted, filtered using human validity, robot validity, and semantic floor criteria, and scored in both human and robot space to select the final humanoid reaction candidate. The example shows a high-five interaction, where the actor is on the right and the reactor is on the left. Candidate 4 is selected over candidate 0 as it preserves the intended interaction while providing more suitable motion references for humanoid retargeting. The visualized G1 motions further illustrate how the final retargeted humanoid motion can be performed in the human--robot interaction scene.}
\label{fig:pipeline}
\vspace*{-5pt}
\end{figure*}
We propose an inference-time search framework that improves the suitability of generated reactions for humanoid robots by searching over multiple outputs of a pretrained reaction generation model while considering embodiment constraints, rather than directly using the first sampled reaction. The key idea is to use the reaction generator as a proposal generator: it first generates diverse candidate reactions, while embodiment-aware criteria determine which candidate provides a more suitable motion reference for the target humanoid without degrading the intended interaction semantics. The complete proposal, retargeting, filtering, and selection pipeline is summarized in \Cref{fig:pipeline}.

We adopt ReMoGen as the base reaction generator and keep all of its parameters frozen at their official checkpoint throughout our experiments. Given the observed interaction context of two people and the inferred interaction condition, the frozen generator produces a set of $K$ candidate reactions for the target person.

We use the causal interaction as the conditioning input to infer the interaction semantics, where ${A}_{1:t}$ is the causal action of the actor and ${R}_{1:t}$ is the causal reaction of the reactor. $\mathcal{S}_{\phi}$ predicts the interaction belief $\mathbf{b}_t$ from the observed history and then maps the most likely interaction class to a text embedding $\mathbf{e}_t$, which is then used to condition the frozen reaction generation model $p_{\theta}$ to generate a set of $K$ candidate reactions $\{\mathbf{y}_{t,k}\}_{k=0}^{K-1}$.

\begin{equation}
\mathbf{b}_t, \mathbf{e}_t
=
\mathcal{S}_{\phi}
\left(
\mathbf{A}_{1:t},
\mathbf{R}_{1:t}
\right),
\end{equation}
\begin{equation}
\mathbf{y}_{t,k}
\sim
p_{\theta}
\left(
\mathbf{y}
\mid
\mathbf{h}_t^{R},
\mathbf{h}_t^{A},
\mathbf{e}_t
\right),
\qquad
k=0,\ldots,K-1.
\end{equation}

We designate the first candidate reaction $\mathbf{y}_{t,0}$ as \emph{candidate 0}. It is sampled from the same generator as other candidates and has no special setting. Candidate 0 serves as the single-sample, no-search baseline: when $K=1$, it is the only generated reaction.

For each candidate reaction, we concatenate the generated reaction $\mathbf{y}_{t,k}$ with the reactor history $\mathbf{h}_t^{R}$ and retarget the complete segment using the frozen retargeting model $\mathcal{T}_{E}$ to obtain the corresponding robot motion $\mathbf{Q}_{t,k}$.
\begin{equation}
\mathbf{Q}_{t,k}
=
\mathcal{T}_{E}
\left(
\mathbf{h}_t^{R}
\oplus
\mathbf{y}_{t,k}
\right),
\label{eq:retargeting}
\end{equation}

To preserve interaction semantics while optimizing embodiment-aware suitability, we introduce the following semantic constraint:

\begin{equation}
s_{t,k}
=
\mathbf{b}_t^{\top}\operatorname{softmax}
\left(
\frac{\boldsymbol{\ell}_{t,k}}{T_s}
\right),
\qquad
s_{t,k} \geq s_{t,0}.
\end{equation}

Here, $\boldsymbol{\ell}_{t,k}$ are the class logits predicted by the frozen semantic recognizer for candidate $k$. The logits are converted into a calibrated class-probability distribution using temperature $T_s$, and their inner product with the interaction-class belief $\mathbf{b}_t$ defines the semantic compatibility score $s_{t,k}$. We retain only candidates whose semantic score is higher than or equal to that of the baseline $k=0$. Candidate reactions that satisfy this constraint are considered eligible for further selection. In our primary configuration, MS-G3D~\cite{liu2020msg3d} is adopted for this inference-time semantic constraint, while CTR-GCN~\cite{chen2021ctrgcn} is kept independent of selection and used as a held-out semantic evaluator.

For each retargeted candidate, we compute a normalized kinematic cost $J_{\mathrm{kin}}(t,k)$ to quantify its suitability for the target embodiment,
\begin{equation}
J_{\mathrm{kin}}(t,k)
=
\sum_{m\in\mathcal{M}_{\mathrm{kin}}}
w_m
\frac{
\max(g_{t,k,m},0)
}{
\sigma_m
},
\label{eq:kinematic_cost}
\end{equation}

where $\mathcal{M}_{\mathrm{kin}}$ denotes the set of robot-side feasibility metrics, $g_{t,k,m}$ is the measured value of metric \(m\) on retargeted candidate $\mathbf{Q}_{t,k}$, and $\sigma_m$ and $w_m$ are the corresponding fixed normalization scale and weight, respectively. The metric set includes retargeting position and end-effector errors, orientation error, maximum joint-limit violation and near-joint-limit fraction, joint-velocity-excess fraction, acceleration and jerk, transition joint jump, maximum self- and floor-penetration depth, and foot sliding. Scales and weights are fixed across experiments.

We additionally define an interaction-preservation cost $J_{\mathrm{int}}(t,k)$ based on hand-distance distortion $d^{\mathrm{hand}}_{t,k}$, hand-interaction timing error $\tau^{\mathrm{hand}}_{t,k}$, and actor--robot minimum distance $d^{\min}_{t,k}$.
\begin{equation}
\begin{aligned}
J_{\mathrm{int}}(t,k)
={}&
w_d
\frac{d^{\mathrm{hand}}_{t,k}}{\sigma_d}
+
w_\tau
\frac{\tau^{\mathrm{hand}}_{t,k}}{\sigma_\tau}
\\
&+
w_c
\frac{
\max\left(0,\,d_{\mathrm{safe}}-d^{\min}_{t,k}\right)
}{
\sigma_c
}.
\end{aligned}
\label{eq:interaction_cost}
\end{equation}
Here, $d_{\mathrm{safe}} - d^{\min}_{t,k}$ penalizes insufficient actor--robot clearance. The coefficients $w_d$, $w_\tau$, and $w_c$ are fixed weights of the three terms, while $\sigma_d$, $\sigma_\tau$, and $\sigma_c$ are normalization scales. In our experiments, we set $w_d=0.45$, $w_\tau=0.45$, $w_c=0.10$, $\sigma_d=0.15$, $\sigma_\tau=0.20$, $\sigma_c=0.03$, and $d_{\mathrm{safe}}=0.03$. Term weights, normalization scales, and thresholds are fixed before evaluation and are kept unchanged across datasets and experiments. No test-set, per-dataset, or per-experiment tuning is performed. Interaction quantities are computed using an observed-only actor forecast, without access to recorded future motion. Specifically, from the actor's last observed pose, we extrapolate its root translation using the mean root translation over recent frames and then apply the same translation to all joints. For robot-space interaction evaluation, the actor forecast is transformed into the robot frame by preserving its relative position with respect to the generated human reactor, replacing the human reactor root with the retargeted robot pelvis as the reference.

Because $J_{\mathrm{kin}}$ and $J_{\mathrm{int}}$ have different numerical scales, we normalize their improvements relative to candidate 0. For a generic cost $J$, we define the normalization factor $\gamma_t(J)$ as follows:
\begin{equation}
\gamma_t(J)
=
\max\left\{
\operatorname{Std}_{j=0}^{K-1} J(t,j),
\;0.01|J(t,0)|,
\;10^{-4}
\right\}.
\label{eq:cost_normalization}
\end{equation}

Using this normalization procedure, we define the standardized kinematic and interaction benefits relative to candidate 0 as $B_{\mathrm{kin}}(t,k)$ and $B_{\mathrm{int}}(t,k)$, respectively.

\begin{equation}
B_{\mathrm{kin}}(t,k)
=
\frac{
J_{\mathrm{kin}}(t,0)
-
J_{\mathrm{kin}}(t,k)
}{
\gamma_t(J_{\mathrm{kin}})
},
\label{eq:kinematic_benefit}
\end{equation}

\begin{equation}
B_{\mathrm{int}}(t,k)
=
\frac{
J_{\mathrm{int}}(t,0)
-
J_{\mathrm{int}}(t,k)
}{
\gamma_t(J_{\mathrm{int}})
}.
\label{eq:interaction_benefit}
\end{equation}

A candidate is considered eligible only if it satisfies the human-motion validity check, robot-motion validity check, and semantic floor constraint,

\begin{equation}
\mathcal{E}_t
=
\left\{
k
\;\middle|\;
h_{t,k}=1,\;
r_{t,k}=1,\;
s_{t,k}\geq s_{t,0}
\right\}.
\label{eq:eligible_set}
\end{equation}

$h_{t,k}\in\{0,1\}$ indicates whether candidate $k$ passes the observable human-motion validity checks. We compute human-space metrics available at inference time, including boundary position, velocity, heading continuity, acceleration, jerk, floor penetration, foot sliding, bone-length consistency, and actor--reactor capsule penetration. Candidate $k$ is human-valid if none of these quantities violates the validity criteria by more than a fixed tolerance. Candidate $k=0$ is retained as a fallback if no other candidate is valid.

$r_{t,k}\in\{0,1\}$ indicates retargeted robot motion validity. It rejects motions with large self-penetration, floor penetration, or joint-limit violation. Using the fixed thresholds,

\begin{equation}
r_{t,k}
=
\mathbb{I}
\left[
d^{\mathrm{self}}_{t,k}
\leq \delta_{\mathrm{self}}
\land
d^{\mathrm{floor}}_{t,k}
\leq \delta_{\mathrm{floor}}
\land
v^{\mathrm{limit}}_{t,k}
\leq \delta_{\mathrm{limit}}
\right],
\label{eq:robot_validity}
\end{equation}

where $d^{\mathrm{self}}_{t,k}$ is the maximum self-penetration depth, $d^{\mathrm{floor}}_{t,k}$ is the maximum floor penetration depth, and $v^{\mathrm{limit}}_{t,k}$ is the maximum joint-limit violation. We use fixed thresholds of $\delta_{\mathrm{self}}=0.03$, $\delta_{\mathrm{floor}}=0.03$, and $\delta_{\mathrm{limit}}=10^{-4}$ across all experiments.

The final candidate $k_t^{*}$ is selected from the eligible set by maximizing the equally weighted kinematic and interaction benefits. If the eligible set is empty, we fall back to the baseline candidate $k=0$.

\begin{equation}
k_t^{*}
=
\arg\max_{k\in\mathcal{E}_t}
\frac{1}{2} \left[
B_{\mathrm{kin}}(t,k)
+
B_{\mathrm{int}}(t,k)
\right].
\label{eq:final_selector}
\end{equation}

Beyond flat independent sampling, this framework can be extended with iterative candidate refinement. After an initial candidate pool is sampled and evaluated, high-ranking candidates are used as centers for random latent perturbations, and the resulting candidates are evaluated using the same criteria. At each subsequent iteration, new refinement centers are selected from the accumulated candidate pool. The final selection is performed over the aggregated candidates from all iterations. We use flat $K=16$ sampling as the primary configuration, while iterative refinement is evaluated separately as a higher-compute search variant.

\section{Experiments}

\subsection{Experimental Setup}
In our experiments, we evaluate the proposed inference-time search framework on the Inter-X benchmark~\cite{xu2024interx}, which provides a diverse set of human--human interactions. We compare our pipeline against the ReMoGen single-sample generation baseline and evaluate both human reaction quality and humanoid retargeting feasibility. In addition, we conduct ablation studies to analyze the contributions of different components in our framework. We also evaluate the inference-time search framework's generalization across different humanoid embodiments, retargeting models, and human interaction datasets.

Unless otherwise specified, the ReMoGen generator is frozen and $K=16$ candidates are sampled at inference time, with candidate 0 serving as the single-sample baseline. GMR and Unitree~G1~\cite{unitree_g1} are used as the primary retargeter and embodiment, respectively. Main comparisons are reported on the official test split, whereas parameter selection, ablations, and diagnostic analyses are performed on the validation split.

\subsection{Results}

\begin{table*}[htbp]
\vspace*{7pt}
\centering
\papertablestyle
\caption{Human reaction quality on Inter-X. (a) Comparison using a semantic condition inferred from the observed interaction history. (b) Validation-only semantic-input diagnostic.}
\label{tab:human_quality}
\begin{tabular*}{\textwidth}{@{\extracolsep{\fill}}llcrrrrrrrrr}
\toprule
\twolinetablecell{Split /}{Semantic Input} & Method & $K$ & FID$\downarrow$ & R@3$\uparrow$ & MM Dist.$\downarrow$ & Diversity$\rightarrow$  & Peak Jerk$\downarrow$ & Collision Proxy$\downarrow$ & Contact F1$\uparrow$ & $\Delta p_y$ (pp)$\uparrow$ \\
\midrule
\multicolumn{11}{l}{\textit{(a) Inter-X test: predicted class from observed history}}\\
Test / Predicted & ReMoGen & 1 & \textbf{1.661} & 0.130 & 12.855 & 6.614 & 0.078 & 10.240 & \textbf{0.770} & 0.00 \\
Test / Predicted & \textbf{Ours} & 16 & 1.737 & \textbf{0.132} & \textbf{12.841} & 6.666 & \textbf{0.072} & \textbf{10.188} & \textbf{0.770} & \textbf{+0.21} [+0.10, +0.32] \\
\midrule
\multicolumn{11}{l}{\textit{(b) Validation-only semantic-input diagnostic}}\\
Val. / Predicted & ReMoGen & 1 & 2.468 & 0.130 & 12.732 & 6.916 & 0.079 & 10.111 & 0.784 & 0.00 \\
Val. / Predicted & Ours & 16 & 2.607 & 0.133 & 12.625 & 6.917 & \textbf{0.074} & \textbf{10.069} & 0.784 & +0.45 \\
Val. / GT class  & ReMoGen & 1 & \textbf{2.229}& 0.140& 12.641 & 6.929 & 0.079 & 10.292 & 0.788 & 0.00 \\
Val. / GT class  & Ours & 16 & 2.446& \textbf{0.141} & \textbf{12.523} & 6.867 & \textbf{0.074} & 10.097 & \textbf{0.791} & \textbf{+0.55} \\
\bottomrule
\end{tabular*}
\vspace*{-7pt}
\end{table*}
We compare our pipeline against the ReMoGen single-sample generation baseline on the Inter-X dataset and evaluate both human reaction quality and humanoid retargeting feasibility. We compare the final selected reactions against the baseline methods in terms of human reaction quality, as shown in Table~\ref{tab:human_quality}. We use ReMoGen's released evaluator for FID, R Precision, MM Distance, Diversity, peak jerk, and contact F1. Because the Collision(\%) metric reported by ReMoGen is not available in its released evaluator, we instead report a collision proxy defined as the percentage of future frames with more than $1\,\mathrm{mm}$ of actor--reactor capsule penetration. $\Delta p_y$ denotes the change in probability assigned to the ground-truth interaction class by a held-out CTR-GCN evaluator, relative to candidate 0. CTR-GCN is not used during candidate selection, therefore it provides an independent measure of reaction semantic preservation. On the validation set, we evaluate performance both with semantics inferred from the causal interaction history and with ground-truth interaction semantics. On the test split, inference-time selection maintains contact F1 while improving R@3, MM distance, peak jerk, collision proxy, and $\Delta p_y$.

\begin{figure}[htbp!]
\centering
\includegraphics[width=0.80\columnwidth]{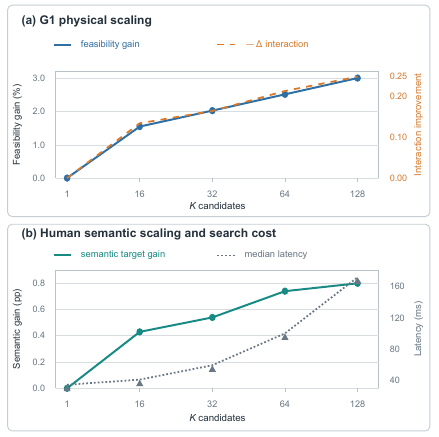}
\caption{Candidate-sampling scaling, with $K=1$ denoting the single-sample no-search baseline. (a) G1 feasibility gain and interaction improvement for the humanoid reaction search space. (b) Semantic gain and median search latency for the human reaction search space.}
\label{fig:k_scaling}
\end{figure}

In Table~\ref{tab:fixed_g1}, we report the fixed-window humanoid retargeting results on the Inter-X dataset. On the official test split, the proposed method achieves a \(+1.61\%\) feasibility gain, with a 95\% confidence interval of [+1.51,+1.72]\%. The proposed inference-time search framework consistently improves robot-side performance, yielding higher retargeting feasibility and robot validity while reducing interaction distortion, foot sliding, and motion jerk, with a slight improvement in semantic score. The ablation study in Panel (b) shows that both semantic and kinematic components contribute to the overall performance, while the final configuration balances robot-side optimization with semantic non-degradation. Removing the semantic floor improves robot feasibility, but eliminates the explicit semantic non-degradation guarantee relative to candidate 0. We therefore retain the semantic floor in the final framework. In addition, the Random human-valid selector, which randomly selects a candidate from the human-valid set, shows no obvious improvement over the baseline, demonstrating the effectiveness of the proposed selection method.

\begin{table*}[htbp]
\vspace*{5pt}
\centering
\papertablestyle
\caption{Fixed-window humanoid retargeting on Inter-X. (a) Comparison on the official test split. (b) Selector component ablation on the validation split.}
\label{tab:fixed_g1}
\begin{tabular*}{\textwidth}{@{\extracolsep{\fill}}llcrrrrrr}
\toprule
Split & Method & $K$ & \twolinetablecell{Feas. Gain}{(\%)$\uparrow$} & \twolinetablecell{$\Delta$}{Interaction$\downarrow$} & \twolinetablecell{$\Delta p_y$}{(pp)$\uparrow$} & \twolinetablecell{Foot}{Slide$\downarrow$} & \twolinetablecell{Joint}{Jerk$\downarrow$} & \twolinetablecell{Robot Valid}{(\%)$\uparrow$} \\
\midrule
\multicolumn{9}{l}{\textit{(a) Official Inter-X test}}\\
Test & ReMoGen single / candidate 0 & 1 & 0.00 & 0.000 & 0.00 & 0.143 & 44.15 & 96.15 \\
Test & \textbf{Ours} & 16 & \textbf{+1.61}[+1.51,+1.72] & \textbf{-0.120} & \textbf{+0.21} & \textbf{0.127} & \textbf{42.01} & \textbf{96.73} \\
\midrule
\multicolumn{9}{l}{\textit{(b) Validation-only selector ablation}}\\
Val. & ReMoGen single / candidate 0 & 1 & +0.00 & +0.000 & +0.00 & 0.142 & 46.81 & 94.48 \\
Val. & Random human-valid & 16 & +0.94 & +0.013 & -0.14  & 0.130 & 43.71 & 94.44 \\
Val. & Semantic-only & 16 & +0.86 & +0.004 & \textbf{+0.43} & 0.131 & 44.54 & 94.41 \\
Val. & Kinematic-only & 16 & \textbf{+3.23}  & +0.016  & -0.18 & \textbf{0.112}& \textbf{42.49} & \textbf{95.46} \\
Val. & Semantic + kinematic & 16 & +2.43  & +0.007  & +0.24 & 0.118 & 43.64 & \textbf{95.46} \\
Val. & Combined, no semantic floor & 16 & +2.12 & \textbf{-0.175}  & +0.19 & 0.120 & 44.03 & \textbf{95.46} \\
Val. & Ours: strict semantic floor + robot ranking & 16 & +1.55 & -0.134 & +0.17 & 0.126 & 44.69 & 95.11 \\
\bottomrule
\end{tabular*}
\vspace*{-5pt}
\end{table*}

\begin{figure}[htbp!]
\centering
\includegraphics[width=\columnwidth]{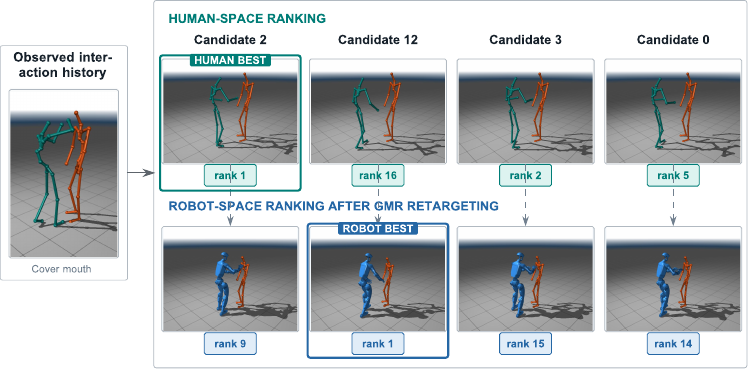}
\caption{Human- and robot-space candidate rankings can be different. The same observed history produces multiple human reaction candidates (top), which are retargeted to the target humanoid (bottom). The highlighted example shows a candidate ranking change due to the retargeting needs.}
\label{fig:space_mismatch}
\end{figure}

Figure~\ref{fig:space_mismatch} illustrates that the candidate preferred in human-motion space does not always remain the most suitable after retargeting. Across matched candidate sets, the human-space and G1 rankings have a mean Spearman correlation of \(\rho=0.366\), and their best-candidate selections disagree in \(78.98\%\) of samples. The limited ranking agreement indicates that human-space reaction quality alone is insufficient to identify the candidate most suitable for the target embodiment.

In Table~\ref{tab:full_rollout}, we report the full autoregressive rollout results on the Inter-X dataset. Panel (a) shows the human reaction quality results, while panel (b) presents the humanoid retargeting results. The \(K=1\) and \(K=16\) human rollout branches are independently retargeted through continuous GMR passes. The proposed inference-time search framework improves both human reaction quality and humanoid retargeting feasibility, while also reducing interaction distortion, foot sliding, and jerk. This demonstrates that the proposed framework not only works on fixed individual windows but also adapts to full sequences. Some of the metrics were not reported for the human reaction results because they are only applicable to robot results.

\begin{table*}[htbp]
\vspace*{5pt}
\centering
\papertablestyle
\caption{Full autoregressive rollout results on the official Inter-X test split.} 
\label{tab:full_rollout}
\begin{tabular*}{\textwidth}{@{\extracolsep{\fill}}llcrrrrrrrr}
\toprule
Space & Method & $K$ & \twolinetablecell{FID$\downarrow$ /}{Feas.$\downarrow$} & \twolinetablecell{MM Dist.$\downarrow$ /}{Gain$\uparrow$} & Interact.$\downarrow$ & Slide$\downarrow$ & Jerk$\downarrow$ & Jump$\downarrow$ & \twolinetablecell{Robot Valid}{(\%)$\uparrow$} \\
\midrule
\multicolumn{10}{l}{\textit{(a) Human reaction quality}}\\
Human & Predicted-class ReMoGen single & 1 & 0.665 & 6.099 & -- & -- & -- & -- & -- \\
Human & \textbf{Predicted-class Ours} & 16 & \textbf{0.587} & \textbf{6.052} & -- & -- & -- & -- & -- \\
\midrule
\multicolumn{10}{l}{\textit{(b) Whole-branch G1 transfer}}\\
Robot & ReMoGen single full branch & 1 & 0.848 & 0.00\% & 3.425 & 0.166 & 59.66 & 0.182  & 80.08 \\
Robot & \textbf{Ours full branch} & 16 & \textbf{0.815} & \textbf{+3.54 [+3.39,+3.69]\%} & 3.408 & \textbf{0.132} & \textbf{53.23} & \textbf{0.166} & \textbf{84.74} \\
\bottomrule
\end{tabular*}
\vspace*{-5pt}
\end{table*}

\subsection{Generalization}

To investigate the generalizability of the proposed inference-time search framework, we evaluate its performance across different retargeters, datasets, and robot morphologies. The results are summarized in Table~\ref{tab:generalization}. In the retargeter generalization setting, we adopt NMR as the secondary retargeter to evaluate the proposed pipeline. We observe that the proposed framework consistently improves feasibility and interaction metrics across both GMR and NMR retargeted robot motions. This also demonstrates that the proposed framework is not limited to a specific retargeting model.

For dataset generalization, we adopt the InterHuman dataset and find that our method outperforms the ReMoGen single-sample baseline on both InterHuman fixed and full sequences, demonstrating its robustness to different interaction contexts and showing that the proposed framework can generalize to different human interaction datasets.

\begin{figure*}[htbp]
  \centering
  \includegraphics[width=0.76\textwidth]{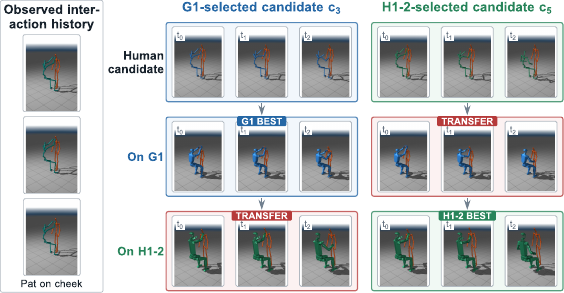}
  \caption{Candidate selection differences between G1 and H1-2. The G1 best and H1-2 best candidates show the optimal retargeted motion for each morphology, and the ``transfer'' block shows the retargeted reaction candidate from the other morphology's optimal candidate.}
  \label{fig:candidate_difference}
\end{figure*}

Finally, in the cross-embodiment morphology dependence setting, our method shows improved performance on both Unitree~G1 and H1-2~\cite{unitree_h1} morphologies, indicating its adaptability to different humanoid embodiments. In particular, the results show that the optimal reaction candidate for different morphologies can be different. Examples of such differences are shown in Figure~\ref{fig:candidate_difference}, where the optimal candidates for different humanoids show variations in pose. In the Inter-X set, the selected candidates for G1 and H1-2 disagree in 38.23\% of the cases, while their candidate rankings have a mean Spearman correlation of \(\rho=0.335\). This suggests that the proposed inference-time search framework can effectively adapt to different robot morphologies by selecting candidates that are more suitable for them.

\begin{table}[htbp]
\centering
\papertablestyle
\caption{Generalization across retargeters, datasets, and robot morphologies. InterHuman dataset results use the official test split. The results are improvements over the ReMoGen single-sample baseline.}
\label{tab:generalization}
\begin{tabular}{lrrrr}
\toprule
Setting & $K$ &Feas. Gain(\%)$\uparrow$ & $\Delta$ Interaction$\downarrow$ & Robot Valid$\uparrow$ \\
\midrule
\multicolumn{5}{l}{\textit{(a) Retargeter generalization}}\\
GMR  & 16 & +1.61 & -0.120 & 96.15$\rightarrow$96.73\% \\
NMR & 16 & +1.24 & -0.099 & 84.65$\rightarrow$85.80\% \\
\midrule
\multicolumn{5}{l}{\textit{(b) Dataset generalization}}\\
InterHuman fixed  & 16 & +2.50 & -0.121 & 80.58$\rightarrow$83.74\% \\
InterHuman full & 16 & +3.85 & -0.034 & 75.84$\rightarrow$79.89\% \\
\midrule
\multicolumn{5}{l}{\textit{(c) Cross-embodiment generalization}}\\
G1 & 16 & +1.61 & -0.120 & 96.15$\rightarrow$96.73\% \\
H1-2  & 16 & +0.86 & -0.116 & 92.10$\rightarrow$93.83\% \\
\bottomrule
\end{tabular}
\end{table}

\begin{table*}[htbp]
\vspace*{5pt}
\centering
\papertablestyle
\caption{Candidate-sampling scaling and matched-budget iterative refinement results.}
\label{tab:scaling_ablation}

\begin{tabular}{llrrrrr}
\toprule
Method &
$K$/Budget &
\twolinetablecell{Human Semantic}{Gain (pp)$\uparrow$} &
G1 Semantic $\Delta$ (pp)$\uparrow$ &
Feas. Gain (\%)$\uparrow$ &
$\Delta$ Interaction$\downarrow$ &
Robot Valid (\%)$\uparrow$ \\
\midrule

\multicolumn{7}{l}{\textit{(a) Flat candidate-sampling scaling}}\\
Flat sampling & 1   & +0.00 & +0.00 & +0.00 & +0.000 & 94.48 \\
Flat sampling & 16  & +0.43 & \textbf{+0.17} & +1.55 & -0.134 & 95.11 \\
Flat sampling & 32  & +0.54 & +0.12 & +2.03 & -0.164 & 95.25 \\
Flat sampling & 64  & +0.74 & +0.16 & +2.52 & -0.213 & 95.35 \\
Flat sampling & 128 & \textbf{+0.80} & +0.09 & \textbf{+3.01} & \textbf{-0.248} & \textbf{95.67} \\

\midrule
\multicolumn{7}{l}{\textit{(b) Matched-budget perturbation and iterative search}}\\
One-stage perturb. &
64 (32+4$\times$8) & +0.74 & +0.20 & +3.19 & -0.223 & 95.67 \\

Two-stage iterative &
64 (16+4$\times$8+2$\times$8) & +0.65 & +0.26 & +3.24 & -0.230 & 95.60 \\

Four-round iterative &
128 (32+4$\times$8+4$\times$8+2$\times$8+2$\times$8) & \textbf{+0.88} & \textbf{+0.28} & \textbf{+4.94} & \textbf{-0.301} & \textbf{95.95} \\

\bottomrule
\end{tabular}
\vspace*{-5pt}
\end{table*}

We then evaluate the effect of the candidate budget $K$ on the performance of the proposed inference-time search framework. The results are summarized in Table~\ref{tab:scaling_ablation} and Figure~\ref{fig:k_scaling}. We observe that increasing the candidate budget improves robot-side feasibility, reduces interaction costs, and increases human-space semantic gain, while also increasing the system latency. However, there are diminishing returns as $K$ increases and a decrease in semantic gain on retargeted G1 motion. $K=16$ therefore provides a balance between computation, feasibility improvement, and semantic preservation.

In addition to flat candidate scaling, we evaluate matched-budget iterative latent-perturbation strategies to investigate whether local refinement-based candidate searching can provide an advantage over independent sampling. Starting from independently sampled ReMoGen candidates, the highest-ranked candidates satisfying the semantic constraint are used as perturbation centers, and perturbations are applied to their ReMoGen latent representations. Previously generated candidates are retained at each stage to preserve promising candidates discovered in each stage, and perturbation centers are selected from the aggregated candidate pool. The one-stage variant starts from 32 independent samples and generates 8 perturbations for each of the top 4 candidates, resulting in 64 candidates in total. The two-stage variant starts from 16 independent samples, adds 8 perturbations to each of the top 4 candidates in the first stage, and then adds 8 perturbations to each of the top 2 candidates selected from the candidate pool, resulting in 64 candidates. We further evaluate a four-round strategy that starts from 32 independent samples, with 8 perturbations applied to the top 4, top 4, top 2, and top 2 candidates, respectively, resulting in 128 candidates in total. Each variant is compared with candidate 0. The results show that iterative refinement consistently improves robot-side optimization over matched-budget flat sampling, indicating that refining promising candidates can exploit additional search headroom. However, the sequential refinement stages increase inference latency, so we retain flat $K=16$ as the primary configuration.

\begin{table}[htbp]
\centering
\papertablestyle
\caption{TWIST tracking performance on the Inter-X test split.}
\label{tab:twist_tracking}
\begin{tabularx}{\columnwidth}{@{}>{\raggedright\arraybackslash}Xrr>{\centering\arraybackslash}p{0.28\columnwidth}@{}}
\toprule
Metric & Baseline & Ours & \twolinetablecell{$\Delta$}{[95\% CI]} \\
\midrule
Tracking success (\%) & 92.6886 & 92.6886 & \twolinetablecell{+0.0000}{[-0.1168,+0.1168]} \\
Joint RMSE (rad) & 0.1782 & 0.1780 & \twolinetablecell{-0.0002}{[-0.0005,+0.0000]} \\
Root position RMSE (m) & 0.0397 & 0.0393 & \twolinetablecell{-0.0004}{[-0.0006,-0.0003]} \\
Root orientation RMSE (rad) & 0.1692 & 0.1678 & \twolinetablecell{-0.0014}{[-0.0021,-0.0009]} \\
Wrist RMSE (m) & 0.0526 & 0.0521 & \twolinetablecell{-0.0005}{[-0.0007,-0.0004]} \\
Torque RMS (Nm) & 13.1829 & 13.1609 & \twolinetablecell{-0.0220}{[-0.0298,-0.0142]} \\
Physics foot sliding (m/s) & 0.103 & 0.100 & \twolinetablecell{-0.002}{[-0.003,-0.002]} \\
Hand-distance distortion (m) & 0.1621 & 0.1602 & \twolinetablecell{-0.0019}{[-0.0026,-0.0012]} \\
Hand-timing error (s) & 0.2486 & 0.2305 & \twolinetablecell{-0.0181}{[-0.0227,-0.0137]} \\
\bottomrule
\end{tabularx}
\end{table}

\begin{table}[htbp]
\vspace*{5pt}
\centering
\papertablestyle
\caption{Runtime of the search and retargeting pipeline (ms).}
\label{tab:runtime}
\begin{tabular}{lrr}
\toprule
Component & Median & p90 \\
\midrule
Human reaction search, $K=16$ & 41.26 & 42.62 \\
GMR, one 10-frame candidate (2 hist. + 8 pred.) & 30.85 & 40.13 \\
Parallel GMR + robot metrics, $K=16$ & 64.81 & 82.72 \\
Complete loaded $K=16$ path & 108.68 & 128.99 \\
\bottomrule
\end{tabular}
\vspace*{-5pt}
\end{table}

\subsection{Practical Evaluation}
For the practical deployment of the proposed inference-time search framework, we evaluate the independent physics tracking and runtime performance. Here we use TWIST~\cite{ze2025twist} as the physics tracking tool to evaluate the retargeted humanoid motion. The results are summarized in Table~\ref{tab:twist_tracking}. We observe that the proposed method maintains comparable tracking success and root RMSE while slightly improving joint RMSE and wrist RMSE. The torque RMS, physics foot sliding, hand-distance distortion, and hand-timing error are also improved. The runtime measurements in Table~\ref{tab:runtime} indicate that the median runtime of the $K=16$ pipeline on a computer equipped with an RTX 4090 GPU and an i9-13900KF CPU is 108.68 ms, corresponding to 9.2 Hz. The 10-frame GMR input consists of 2 observed history frames and 8 predicted future frames. The measured runtime is therefore compatible with near-real-time inference over the 8-frame prediction horizon. The complete loaded $K=16$ path ends at the selected G1 motion without actual actuation.

\section{Conclusion}
In this work, we propose an inference-time search framework for embodiment-aware human reaction selection in humanoid retargeting. The proposed pipeline samples multiple candidate reactions, retargets them to the target humanoid, and selects among eligible candidates using interaction-preservation and embodiment-specific feasibility criteria. Experimental results show that our approach improves humanoid retargeting feasibility while maintaining overall human-reaction quality. We further find that the reaction preferred in human-motion space can differ from the candidate most suitable after humanoid retargeting, motivating embodiment-aware reaction selection. Results also demonstrate that the proposed framework generalizes across different retargeters, human--human interaction datasets, and humanoid morphologies.

\noindent
{\bf Limitations and Future Work} Despite its optimization gains, the proposed approach requires substantial inference-time computation, particularly when deeper iterative refinement is used. Future work will explore more efficient embodiment-aware candidate generation methods, as well as real-time deployment on physical humanoid robots.

\bibliographystyle{IEEEtran}
\bibliography{main}

\end{document}